\documentclass{article}
\usepackage{spconf,graphicx}
\usepackage[fleqn]{amsmath}
\usepackage{hyperref}

\usepackage{enumitem}
\setlist{nosep, leftmargin=14pt}

\usepackage{mwe} 

\title{Semi-supervised Medical Image Segmentation Method Based on Cross-pseudo Labeling Leveraging Strong and Weak Data Augmentation Strategies
}
\name{
\parbox{\linewidth}{\centering
Yifei Chen$^1$, Chenyan Zhang$^1$, Yifan Ke$^1$, Yiyu Huang$^1$, Xuezhou Dai$^1$, \\
Feiwei Qin$^{1,\star}$, Yongquan Zhang$^2$, Xiaodong Zhang$^{3,4}$, Changmiao Wang$^{5,\star}$}
}

\address{$^1$Hangzhou Dianzi University, Hangzhou, China \\
$^2$Zhejiang University of Finance and Economics, Hangzhou, China \\
$^3$Shenzhen Institute of Advanced Technology, Chinese Academy of Sciences, Shenzhen, China \\
$^4$Shenzhen Children's Hospital, Shenzhen, China \\
$^5$Shenzhen Research Institute of Big Data, Shenzhen, China \\
\tt$\!\!\!\star$ Corresponding Author: qinfeiwei@hdu.edu.cn,cmwangalbert@gmail.com\\}

\begin{document}


\maketitle

\begin{abstract}
Traditional supervised learning methods have historically encountered certain constraints in medical image segmentation due to the challenging collection process, high labeling cost, low signal-to-noise ratio, and complex features characterizing biomedical images. This paper proposes a semi-supervised model, DFCPS, which innovatively incorporates the Fixmatch concept. This significantly enhances the model's performance and generalizability through data augmentation processing, employing varied strategies for unlabeled data. Concurrently, the model design gives appropriate emphasis to the generation, filtration, and refinement processes of pseudo-labels. The novel concept of cross-pseudo-supervision is introduced, integrating consistency learning with self-training. This enables the model to fully leverage pseudo-labels from multiple perspectives, thereby enhancing training diversity. The DFCPS model is compared with both baseline and advanced models using the publicly accessible Kvasir-SEG dataset. Across all four subdivisions containing different proportions of unlabeled data, our model consistently exhibits superior performance. Our source code is available at \href{https://github.com/JustlfC03/DFCPS}{https://github.com/JustlfC03/DFCPS}.
\end{abstract}
\section{Introduction}
\label{sec:intro}

In clinical medicine, medical imaging techniques including DR, CT, and MRI are frequently employed for comprehensive patient diagnosis. These images are then reviewed by medical professionals to identify potential lesions in a patient's internal organs. However, this process is time-consuming, costly, and can lead to physician fatigue, which may compromise diagnostic accuracy. Therefore, medical image segmentation technology has emerged as a critical tool to enhance efficiency and precision in medical consultations \cite{c17liu2021review}. However, unlike segmentation tasks in natural images, medical images often present unique challenges such as low light intensity, low signal-to-noise ratio, and poor contrast. Additional variables such as organ deformation and individual variability can further complicate algorithm design \cite{c14panayides2020ai}. The distribution and sharing of medical images are also legally and ethically regulated due to the sensitive content they contain, including detailed depictions of patient body parts and condition analyses \cite{c15kaissis2020secure}. This often leads to the availability of only undersized datasets. 
To overcome these limitations, semi-supervised learning has attracted considerable attention as a potential method for medical image segmentation.

The Cross Probability Consistency (CPC) model is a streamlined adaptation of the GCT model \cite{c22020guided}. Retaining the core structure of the GCT model, it effectively encapsulates its underlying principles despite its simplified format. This model utilizes consistency learning to engage collaboratively with unlabeled data, deploying two new constraints that function independently of task-specific attributes. Additionally, we present the Cross Pseudo Supervision (CPS) model, which incorporates cross-pseudo labeling as proposed by Chen X et al \cite{c4chen2021semi}. This method leverages consistency learning to stimulate two perturbation networks, facilitating the generation of remarkably similar prediction results for identical input images. Concurrently, unlabeled data augmented with pseudo-labels are employed to expand the training dataset, implementing a self-training mechanism that effectively addresses the insufficiency of labeled data. Moreover, Fixmatch \cite{c5sohn2020fixmatch} is a semi-supervised method proposed by Google Brain and others, which addresses the scarcity of labeled data in comprehensive methods. Distinguishing itself from previous methodologies, FixMatch harnesses cross-entropy to compare weakly and strongly augmented unlabeled data, and has displayed promising results.

Inspired by previous research, we propose an innovative semi-supervised neural network design method, Dual Fixmatch Cross Pseudo Supervision (DFCPS). Building upon and extending the Fixmatch \cite{c5sohn2020fixmatch} concept, this method effectively leverages unlabeled data, a minimal volume of labeled data, and both strong and weak data augmentation techniques to enhance the model's performance. Furthermore, we contrast 
our DFCPS model with baseline and advanced models using the Kvasir-SEG \cite{c6jha2020kvasir} dataset. Our proposed model consistently outshines the others, thereby underscoring its superior performance.

\section{Methods Design}

\subsection{Overview of the Model}

\begin{figure*}[htb]
\centerline{\includegraphics[width=0.9\textwidth]{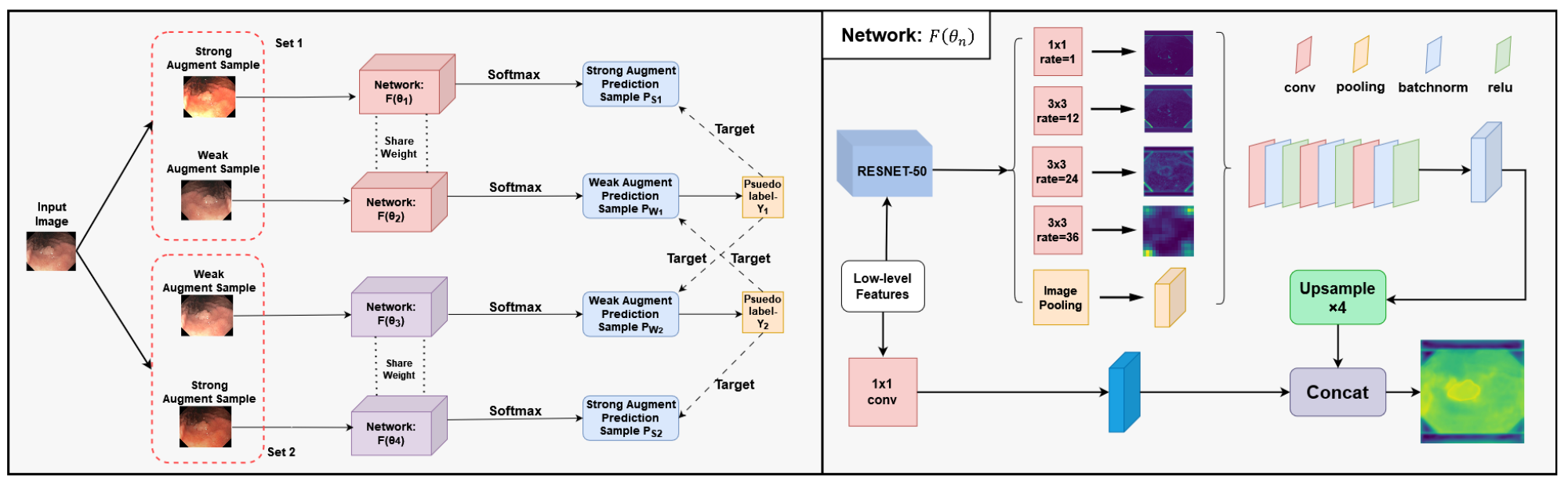}}
\caption{The left half is the overall structure of the DFCPS model. This method incorporates a semi-supervised framework design by combining a cross-pseudo-labelling strategy with a strong and weak data enhancement strategy. The right half is the specific structure of $F(\theta_n)$ neural network. The low level feature is the weakly enhanced feature of the unlabeled sample.}
\label{fig1}
\end{figure*}

The comprehensive structure of the DFCPS model is presented in Fig. \ref{fig1} left half. Initially, both strong and weak augmentation treatments are applied to the same unlabeled original sample, each exhibiting varying degrees of augmentation. These two sets of samples are independently fed into four distinct neural networks, $F(\theta_n)$, for training. Within each group, two neural networks share model parameters and weights, facilitating efficient parameter learning.

The pseudo-label $Y_n$, generated from the prediction result $P_w$ derived from the weakly augmented branch in each group, serves as the target for the prediction of the strongly augmented samples in each instance. Moreover, to bolster the model's robustness and consistency, the concept of cross-pseudo-supervision is integrated into the model design. The pseudo-labels generated from different groups of weakly supervised samples are mutually constrained to enhance the overall consistency of the segmentation results. This constraint mechanism aids in mitigating the impact of noise in the labeled data, thereby further improving the model's stability and robustness.

\subsection{Specific Design}

The model architecture designed for this study employs a coherent learning approach, which we will illustrate by proceeding through a set of strongly and weakly augmented samples. Initially, as depicted in Fig. \ref{fig1} right half, unlabeled image samples undergo weak augmentation such as random rotation, horizontal translation, and are subsequently fed into the Resnet-50 backbone network. Following the generation of the corresponding outputs, the samples enter the Atrous Spatial Pyramid Pooling (ASPP) module, which draws from the DeepLabv2 \cite{c16chen2017deeplab}. This module facilitates multi-scale feature extraction through the use of multiple parallel branches, and enhances the receptive field via atrous convolution. ASPP module enlarges the receptive field without increasing model computational effort, thus improving performance while preserving computational efficiency.
We additionally implemented batch normalization on each small batch of data to mitigate the impact of internal covariate shift, which ultimately leads to a more stable distribution of input features. This not only reduces the potential for vanishing or exploding gradients, but also enhances the overall stability of the network.


An enriched feature representation is realized by concatenating the outcomes from all parallel atrous convolutional and spatial pyramid pooling branches. This is followed by sequential steps of convolution, pooling, batch normalization, and ReLU activation functions. The resulting composite feature is then subjected to an up-sampling technique to enhance data details, thereby improving image quality. These up-sampled composite features are amalgamated with the convolved base features from the weakly augmented branch, thus forming a higher-level feature representation. The model's output is then mapped, pixel-wise, to the probability of each category using the Softmax function. The category with the highest probability is selected to generate predicted samples. Since this study emphasizes binary classification, the model can generate segmentation effects to produce pseudo-labels.

The strongly augmented samples within the same group generate corresponding prediction samples through the same neural network structure. The prediction samples generated by the weakly augmented samples serve as their targets for supervised learning, facilitated by the $L_S$ loss function. Moreover, cross-pseudo-supervised learning is implemented between the prediction samples generated by the weakly augmented samples across the two groups, with the resultant loss determined by the $L_{CPS}$ loss function.

\subsection{Loss Function}

The training regimen for the comprehensive neural network hinges on two primary loss functions: the supervised loss function, denoted as $L_S$, and the cross-pseudo-supervised loss function, represented as $L_{CPS}$. Here, $D^l$ signifies the original set of labeled samples while $D^u$ corresponds to the original set of unlabeled samples. The area of the input image is defined as S, and it's computed using the product of height and width (H*W). The confidence vectors are denoted by $p_i$ and $p_j$. The standard label ground truths are represented as $y_i$ and $y_j$, and the $l_{ce}$ is cross-entropy loss function.

\begin{figure}[htbp]
\centerline{\includegraphics[width=0.4\textwidth]{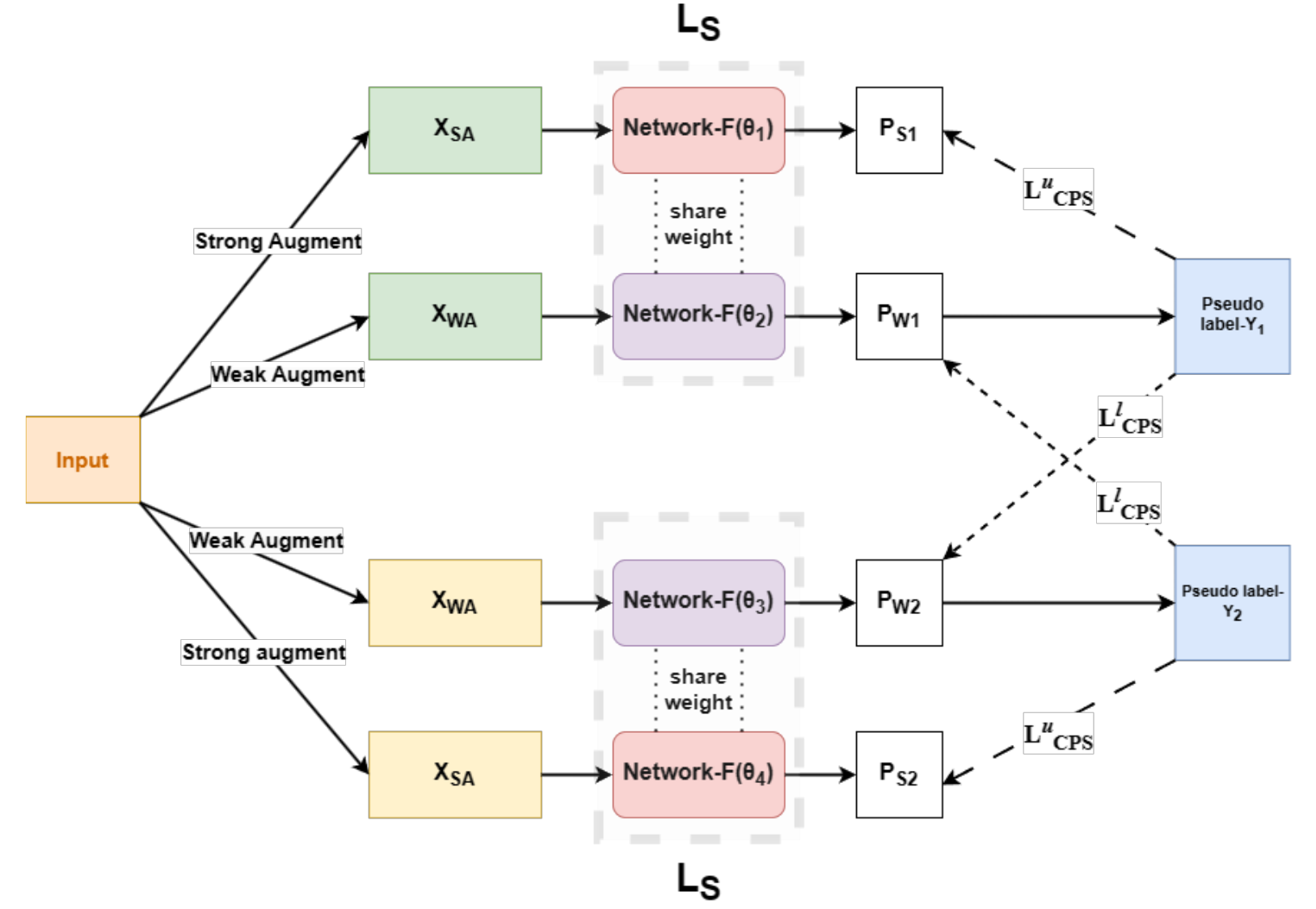}}
\caption{Loss function design of the DFCPS model. The overall architecture of the DFCPS model involves two key loss functions: the supervised loss function $L_S$ and the cross-pseudo-supervised loss function $L_{CPS}$.}
\label{fig3}
\end{figure}

The supervisory loss, denoted by $L_S$, is dictated by the cross-entropy loss function, which executes standard supervised learning on the initial labeled samples. The expression for the supervisory loss function is defined as equation (\ref{eq1}):

\vspace{-1.0em}
\begin{equation}
L_S=\frac{1}{|D^l|}\sum\frac{1}{S}\sum_{x\in{D^l}}^{S}(l_{ce}(p_i,y_i)+l_{ce}(p_j,y_j)).
\label{eq1}
\end{equation}
\vspace{-1.0em}

For the original unlabeled samples, the cross-pseudo-supervision loss is defined as $L_{CPS}^u$ and $L_{CPS}^l$, as conveyed in equations (\ref{eq2}) and (\ref{eq3}). The $L_{CPS}^u$ component employs the pseudo-labels generated by weakly-enhanced samples as the targets of strongly-enhanced samples. It then evaluates the difference between the predictions produced by the strongly-enhanced samples after neural network processing and the corresponding weakly-enhanced samples' pseudo-label. This approach facilitates the network's learning of the mapping relationship from weakly enhanced samples to strongly enhanced ones, thereby enhancing the prediction accuracy of the model on strongly enhanced samples. The $L_{CPS}^l$ component treats the pseudo-labels generated by each batch of weakly enhanced samples as targets for other batches, which aids in constraining the discrepancies between the pseudo-segmentation maps generated from various batches. This technique also encourages interaction and correction among pseudo-labels from different groups, thereby improving the overall segmentation consistency of semi-supervised model. Hence, the total target loss for the entire training procedure can be subsequently defined as per equation (\ref{eq4}).

\vspace{-1.5em}
\setlength{\mathindent}{0cm} 
\begin{gather}
L_{CPS}^u=\frac{1}{|D^u|}\sum\frac{1}{S}\sum_{x\in{D^u}}^S(l_{ce}(P_{s1},Y_1)+l_{ce}(P_{s2},Y_2)),\label{eq2}\\
L_{CPS}^l=\frac{1}{|D^u|}\sum\frac{1}{S}\sum_{x\in{D^u}}^S(l_{ce}(P_{w1},Y_2)+l_{ce}(P_{w2},Y_1)),\label{eq3}\\
Loss=L_S+\omega(L_{CPS}^l+L_{CPS}^u).\label{eq4}
\end{gather}
\vspace{-1.5em}

To ensure a high level of quality in the pseudo-label, this research integrates the notion of a confidence threshold. Pseudo-labels, whose confidence levels align closely with the threshold, are disregarded due to potential reliability concerns. Conversely, those with confidence levels exceeding the threshold are included in the loss calculation.

\section{Results and Discussion}
\subsection{Datasets}


The Kvasir \cite{c11pogorelov2017kvasir} dataset represents the inaugural multi-category dataset designed for the detection and categorization of gastrointestinal (GI) diseases, covers endoscopic images of a wide range of common GI diseases, such as polyps and ulcers. Each image is furnished with pixel-level segmentation labels that demarcate the boundaries of various lesion regions within the image. For the experiments conducted in this study, we employ the Kvasir-SEG \cite{c6jha2020kvasir} segmented polyp dataset, which enhances the dataset's quality by replacing the 13 images from the polyp category in the original dataset with new images. We utilized a random selection of 1,000 intestinal polyp images from the Kvasir-SEG \cite{c6jha2020kvasir} dataset. According to the division protocol of the CPC  \cite{c22020guided} model, the dataset is divided into two groups according to random selection: one group contains 1/2, 1/4, 1/8, and 1/16 of the labeled data, while the other group is the rest of the unlabeled data is called the unlabeled group, which is used to simulate the label-scarcity scenario.

\subsection{Experimental Details}

The experimental server employed in this study is outfitted with six Nvidia GTX 2080Ti graphics cards, each boasting a memory size of 64GB. The initial phase of our experiment entailed pre-training on the PASCAL VOC 2012 dataset \cite{c3everingham2015pascal}. Subsequently, the model was fine-tuned, concentrating on a specific task within the Kvasir-SEG dataset \cite{c6jha2020kvasir}. During the pre-training phase on the PASCAL VOC 2012 \cite{c3everingham2015pascal}, we trained the model for 60 cycles, utilizing a base learning rate of 0.01. Following this, we transferred the pre-trained model weights to the Kvasir-SEG \cite{c6jha2020kvasir} and proceeded with training for an additional 100 epochs. The batch size was set to 12 by default, while the learning rate was adaptively adjusted, with a maximum value of 1e-4 and a minimum value of 1e-6.

\subsection{Comparative Experiment}

The CPC \cite{c22020guided} and CPS \cite{c4chen2021semi} models were employed as baselines for evaluation purposes, alongside our designed model, to assess its capabilities. During the evaluation phase, the same backbone network was utilized for training and testing both the baseline models and our designed model. Furthermore, Table \ref{tab1} displays the mIoU values of DFCPS in comparison to the two baseline models, as well as the state-of-the-art methods ELN \cite{c12kwon2022semi} and ACL-Net \cite{c13wu2023acl}, across different labeling scales of datasets. The RESNET-50 model was chosen as the basis for the backbone network. Notably, DFCPS showcased superior performance across all data scales.





\vspace{-1.5em}
\begin{table}[htb]
\caption{Comparison of mIoU values of baseline and state of the art semi-supervised network models.}
\vspace{-1em}
\begin{center}
\setlength{\tabcolsep}{3.8mm}{
\begin{tabular}{ccccc}
\hline
Methods & 1/2 & 1/4 & 1/8 & 1/16 \\
\hline
CPC \cite{c22020guided} & 77.91 & 76.10 & 73.01 & 67.36 \\
CPS \cite{c4chen2021semi} & 78.47 & 76.74 & 75.66 & 70.50 \\
ELN \cite{c12kwon2022semi} & 75.23 & 73.14 & 71.19 & 71.12 \\
ACL-Net \cite{c13wu2023acl} & 80.07 & 76.94 & 74.83 & 71.27 \\
\textbf{DFCPS(Ours)} & \textbf{80.12} & \textbf{77.42} & \textbf{76.53} & \textbf{72.39} \\
\hline
\end{tabular}}
\label{tab1}
\end{center}
\end{table}
\vspace{-1.5em}

In addition, we also compared the DFCPS model with the two baseline and the state of the art models in terms of training time in the experimental setting of Six-card 2080ti. As shown in Table \ref{tab2}, the training time of DFCPS rises slightly compared to the CPC \cite{c22020guided} and CPS \cite{c4chen2021semi} models. This is because the two baseline models omit the steps of feature consistency loss and backpropagation, and thus the training time may be relatively short. However, the DFCPS model is the least time consuming when reasoning after completing the training, and the results show much better results, and we believe that such an exchange is valuable.

\vspace{-1.5em}
\begin{table}[htb]
\caption{Comparison of training and inference times of baseline and state of the art semi-supervised network models.}
\vspace{-1em}
\begin{center}
\begin{tabular}{p{0.3\linewidth}p{0.27\linewidth}p{0.27\linewidth}}
\hline
Methods & Training Time (hours/epoch) &  Inference Time (per/image) \\
\hline
CPC \cite{c22020guided} & 4.7 & 2.60 \\
CPS \cite{c4chen2021semi} & 5.1 & 2.44 \\
ELN \cite{c12kwon2022semi} & 5.7 & 2.71 \\
ACL-Net \cite{c13wu2023acl} & 5.9 & 2.53 \\
\textbf{DFCPS(Ours)} & \textbf{5.3} & \textbf{2.37} \\
\hline
\end{tabular}
\label{tab2}
\end{center}
\end{table}
\vspace{-1.5em}

\subsection{Ablation Experiment}

In our ablation experiments, we examined the impact of various enhancement strategies on the performance of the model. The outcomes are presented in Table \ref{tab3}, revealing that the strong-weak enhancement combination we implemented yielded the highest mIoU values across different labeling ratio datasets. This observation indicates the reasonability of the data enhancement strategies employed in this study. Furthermore, it is worth noting that the weak-weak enhancement combination outperformed the strong-strong enhancement combination. Most notably, the model that directly employed the original sample images without any data enhancement strategy demonstrated the lowest performance.

\vspace{-1.5em}
\begin{table}[htb]
\caption{Comparison of mIoU values under different data enhancement strategies.}
\vspace{-1em}
\begin{center}
\begin{tabular}{ccccc}
\hline
Enhancement combination & 1/2 & 1/4 & 1/8 & 1/16 \\
\hline
\textbf{strong-weak} & \textbf{80.12} & \textbf{77.42} & \textbf{76.53} & \textbf{72.39} \\
weak-weak & 79.75 & 77.28 & 76.45 & 71.77 \\
strong-strong & 79.63 & 77.04 & 76.28 & 71.23 \\
original sample & 78.47 & 76.74 & 75.66 & 70.50 \\
\hline
\end{tabular}
\label{tab3}
\end{center}
\end{table}
\vspace{-1.5em}

\section{Conclusion}



The DFCPS model, a semi-supervised neural network designed for medical image segmentation, has been carefully devised to encompass the entire process of pseudo-label generation, filtering, and refining. Notably, we have introduced the concept of cross-pseudo-supervision, which marries coherent learning with self-training. This approach empowers the model to harness pseudo-labels from multiple angles, thereby effectively exploiting the unlabeled data to bolster both performance and generalization capabilities. Furthermore, by employing suitable pseudo-labeling generation strategies and filtering mechanisms, we significantly enhance not only the quality and accuracy of the pseudo-labels, but also the performance and robustness of the model.

\section{Compliance with Ethical Standards}
This research study was conducted retrospectively using human subject data made available in open access by Kvasir-SEG \cite{c6jha2020kvasir} dataset. Ethical approval was not required as confirmed by the license attached with the open access data.

\section{Acknowledgment}
This work was supported by Natural Science Foundation of Zhejiang Province (No. LY21F020015), National Natural Science Foundation of China (No. 61972121) , the Open Project Program of the State Key Laboratory of CAD\&CG (No. A2304), Zhejiang University, GuangDong Basic and Applied Basic Research Foundation (No. 2022A1515110570), Innovation Teams of Youth Innovation in Science, Technology of High Education Institutions of Shandong Province (No. 2021KJ088) and National College Student Innovation and Entrepreneurship Training Program (No. 202310336074).




\end{document}